# Machine Learning Based Object Tracking

Md Rakibul Karim Akanda, Joshua Reynolds, Treylin Jackson, and Milijah Gray

Department of Engineering Technology, Savannah State University—Savannah, GA 31404, United States of America

**Abstract**

Machine learning based object detection as well as tracking that object have been performed in this paper. The authors were able to set a range of interest (ROI) around an object using Open Computer Vision, better known as OpenCV. Next a tracking algorithm has been used to maintain tracking on an object while simultaneously operating two servo motors to keep the object centered in the frame. Detailed procedure and code are included in this paper.

1. Introduction

The Raspberry Pi is still selling for a large premium due to the worldwide supply and demand issues. At the time of this report there are still no primary vendors that that had a Raspberry Pi 4b 8gb. We started looking around at the Raspberry Pi alternatives. There are two manufacturers that stand out on paper. The first was a company that makes a product called the Orange Pi, the second is a company that makes a product called the Rock Pi, they have recently dropped the Pi from their name. After purchasing the Rock 4c+ we realized that it and the Raspberry Pi were greatly different in the software side and compatibility of things. This forced us to the secondary market where the prices were at least triple what the MSRP was.

After purchasing the Raspberry Pi 4b 8gb, cooling case, monitor and cables needed. We got to work on installing software, starting to look at code and research the best means to bring our project to fruition. Our first snag came when just trying to install the necessary software, luckily one of the reasons we bit the bullet and paid the money for a Raspberry Pi was that the community is robust. You can find help tutorials and software across the Internet. Unfortunately, that also meant an information overload, we would start installing a set of software which is referred to as packets. These packets wouldn't always be compatible with other packets that we installed and expected them to work with. We would get going on something and find that the time frame of the information was already outdated, and the new software versions wouldn't work with other software versions anymore. Sometimes having the latest and greatest version wouldn't always be beneficial as there wouldn't be a lot of information on it or the workarounds weren't there for bugs that were encountered. It took some trial and error to find the editions that we decided to go with.

One method of safeguarding is to create virtual environments within the folder structure of the Raspberry Pi. This creates a place where you can install different software versions and not have it affect or corrupt the base Raspberry Pi operating system, which was called Raspbian, and its stability. You could essentially Muck up a whole virtual build and instead of starting from scratch and losing all the information on your Pi, such as your codes, you could just delete the virtual environment and create a new one.

In the beginning we started off with the Raspbian 32-bit OS, we decided after encountering so many issues with packet compatibility and software execution stability that we would start with an OS base that had been tried and tested with plenty of resources and support. Raspberry Pi has recently come out with a 64-bit operating system, which it renamed Raspberry Pi OS, about a year ago. We were worried however that the bugs would not have been worked out yet. We would later re-examine that after learning how much faster a 64-bit OS would be vs its predecessor and find some good resources. Most information we examined put the 64-bit OS with our Pi at about a 25% performance increase.

Our software selection took a little bit of trial and error, before and after the OS swap. The software we decided to use to execute our plan is listed in the section previous. By the time we decided to go with a 64-bit version we had become pretty good at using the Raspberry Pi system and community to find solutions to our software issues. The 64-bit system came with some fundamental packet changes that affected the Python operation code we developed. The camera had gone from the legacy raspicam software packets to a newer packet system called libcamera which featured a code called picamera2. This change in the packet library caused us to make some changes to the fundamental code that grabs and shows images from the camera sensor.

The physical wiring of the whole system and installing it on the case is all simple there is a nice mounting system in the case with cutouts for all the I/O on the raspberry pi the case came with a nice adapter that allows the 40 pin General Purpose Input Output (GPIO) pins to be accessed on the side. This 40-pin diversion adapter gave us the ability to mount the pan tilt Hat on the side of the case, as the hats take up all 40 pins. This Hat board also comes with pins that can still utilize the Pi's unused 40 pin GPIO. The head of the pan tilt camera has two servos mounted perpendicular on X and Y axes. The system has two cables to operate the servo motors, one cable to connect the camera, fan power cable, a USB-C power cable, and a display cable for the monitor. We went with a USB Bluetooth keyboard and mouse and the system sits on a wooden frame.

The code that we developed after some grueling trial and error turned out to be straight forward. It will be listed in the supplements at the end with notations. The basics of the code are to start the camera capture a frame, draw a box over a range of interest (ROI), use a built-in algorithm to track ROI, start a loop that captures frames, determines the center of the frame determines the center of the ROI, takes the error, and adjust the servos accordingly to minimize that error. Thus, keeping the ROI in the center of the frame. This process will have further detail in the procedures.

2. **Instruments and components and Software:**
- Raspberry Pi: Model 4b
    - Quad Core Cortex-A72(ARM v8) 64-bit @1.8GHz
    - 8 Gb RAM
    - Dual Band IEEE 802.11ac wireless
    - Bluetooth 5.0
    - Gigabit Ethernet
    - 2 x USB 3.0 2 USB 2.0 ports
    - RPi standard 40 Pin GPIO header
    - 2 x micro-HDMI ports support up to 4k @ 60 Hz
    - 2 Lane MIPI DSI
    - 2 Lane MIPI CSI
    - 5v USB-C power connector (min 3A)
- Case
    - Cooling Fan
    - Heat sinks
        - CPU
        - RAM
        - Graphics chip

- Mouse and keyboard
- Monitor
- Sd card 64Gb (> 32 Gb or more recommended)
- Sun founder Pan Tilt camera Set
  - Frame
  - Pan-tilt Hat
  - 2 x Servo Motors
  - Camera Module (5Mp)
  - FCC Cable (MIPI cam cables)
- Micro HDMI to HDMI Cable
- Micro HDMI to MINI HDMI Cable
- Raspberry Pi compatible Switching USB-C 5v 3.6A Power Supply
- Traveling monitor
  - Power Supply
- Raspberry "bullseye" 64 bit with desktop v. 5.15
- Python 3.9.2
- Open Computer Vision 4.7.2
- Thonny (Python Integrated Development Environment)
- Numpy v. 1.19.2

3. **Procedure**

Here importing servo library and other libraries are needed at the beginning of the program. After that we added some more setup code here, we declare dimension **dispH** and **dispW** which we will call up later for the frame. Also, we added variables and initial starting values for your servo motors. After the camera configuration we added the tracker variable and attached the OpenCV code to it. We added the code for capturing a single frame. Next, we declared a bounding box variable and have OpenCV bring up a window to select a Range of Interest (ROI). Then just below that we initialized the tracker and kill the selector window. Back in the while loop just after the image grab, we added a line that checks for a successful range of interest in the tracker. After that we tell the program to assign **x,y,w,h** variables to the associated points within the bounding box for x, y starting coordinate and the width and height of the frame. Next, we add code that will declare an error between the center of the bounding box and the center of the frame then use this error to make the pan and tilt servo motors move to minimize the error and try to move the center of the bounding box to the center of the frame. If the error is less than 40 pixels on the x axis and less than 20 on the y axis the camera doesn't move. If the error drifts farther than that it will move one degree, if it drifts too far then it will adjust by 3 degrees.

4. **Camera tracking code with CSRT tracking:**

```
import cv2 #open computer vision
import time #time functions
from servo import Servo #servo library that converts code PWM frequency to angles
import picamera2 # library that intercats with camera

fps = 0
fpsPosition = (30,30) #where it is displayed
fpsFont = cv2.FONT_HERSHEY_SIMPLEX #font to be used
fpsColor = (255,120,200) #color in BGR
fpsHeight = 0.5 #Font Height
fpsFontThickness = 2 #font thickness

dispH = 800  #going to be the Height of the frame in pixels
dispW = 1000  #going to be the Width of the frame in pixels

pan = Servo(pin = 13) #pin servo signal wire is attached to
tilt = Servo(pin = 12) #pin servo signal wire is attached to
panAngle = 0 #starting position positive is CCW
tiltAngle = -20 #starting point
pan.set_angle(panAngle) #executes starting angle
tilt.set_angle(tiltAngle) #executes starting angle

picam2 = picamera2.Picamera2()
picam2.preview_configuration.main.size = (dispW,dispH) #Configures display window
picam2.preview_configuration.main.format = "RGB888"  #configures display format
picam2.preview_configuration.controls.FrameRate=15   #FPS it strives for
picam2.preview_configuration.align() #aligns frame array
picam2.configure("preview") #applies configurations
picam2.start() #starts camera

tracker=cv2.TrackerCSRT_create() #Tracking code from cv2 library set to a tracker label

img= cv2.flip(picam2.capture_array(),-1)  #captures a single frame

bb=cv2.selectROI(img) #calls for creation of a bounding box and places it to bb variable
```

```python
tracker.init(img,bb)  #initiates tracker from above

cv2.destroyWindow('ROI selector')  #destroys ROI selector Window

while True:
    tstart = time.time() #takes a time stamp of start of loop

    img= cv2.flip(picam2.capture_array(),-1) #grab image and flips it due to camera orientation

    success,box=tracker.update(img) #checks to see if cv2 the was successful in finding an array to track
    if success: #if there is a tracker array
        (x,y,w,h)=[int(a)for a in box] #assigns x,y,w,h to the attributes of the bounding box
        cv2.rectangle(img,(x,y),(x+w,y+h),(255,0,0),2) #tells opencv to draw this rectangle around the bb dimensions

        errorx = (x+w/2)-dispW/2 #creates a variable errorx that is the distance in pixels from the
                                 center of frame and center of bounding box width
        errory = (y+h/2)-dispH/2 #creates a variable errorx that is the distance in pixels from the
                                 center of frame and center of bounding box height

        if errorx > 40:      #creating an "okay"range of error within 40 pixels or it pans CCW (left) by 1 degree
            panAngle = panAngle - 1
            if panAngle < -90:   #till it gets to 90 degrees CCW
                panAngle = -90
            pan.set_angle(panAngle)
        if errorx < -40:     #creating an "okay"range of error within 40 pixels or it pans CW (right) by 1 degree
            panAngle = panAngle + 1
            if panAngle > 90:    #till it gets to 90 degrees CW
                panAngle = 90
            pan.set_angle(panAngle)
        if errorx > 120:     #creating an "okay"range of error within 120 pixels or it pans CCW (left) by 3 degree
            panAngle = panAngle - 3
            if panAngle < -90:   #till it gets to 90 degrees CCW
                panAngle = -90
            pan.set_angle(panAngle)
        if errorx < -120:    #creating an "okay"range of error within 40 pixels or it pans CW (right) by 3 degre
            panAngle = panAngle + 3
            if panAngle > 90:    #till it gets to 90 degrees CW
                panAngle = 90
            pan.set_angle(panAngle)
```

```python
        if errory > 20:       #creating an "okay"range of error within 20 pixels or it tilts down by 1 degree
            tiltAngle = tiltAngle + 1
            if tiltAngle < -40:   #till it gets to 40 degrees
                tiltAngle = -40
            tilt.set_angle(tiltAngle)
        if errory < -20:      #creating an "okay"range of error within 20 pixels or it tilts up by 1 degree
            tiltAngle = tiltAngle - 1
            if tiltAngle > 90:    #till it gets to 90 degrees
                tiltAngle = 90
            tilt.set_angle(tiltAngle)
        if errory > 60:       #creating an "okay"range of error within 60 pixels or it tilts down by 3 degree
            tiltAngle = tiltAngle + 3
            if tiltAngle < -40:   #till it gets to 40 degrees
                tiltAngle = -40
            tilt.set_angle(tiltAngle)
        if errory < -60:      #creating an "okay"range of error within 60 pixels or it tilts up by 3 degree
            tiltAngle = tiltAngle - 3
            if tiltAngle > 90: #till it gets to 90 degrees
                tiltAngle = 90
            tilt.set_angle(tiltAngle)

    cv2.putText(img,str(int(fps))+" FPS",fpsPosition,fpsFont,fpsHeight,fpsColor,fpsFontThickness)#display FPS
    cv2.imshow("Track",img) #show image frame
    tend = time.time() #ends time for FPS
    looptime = tend-tstart #calculates how long one loop takes

    fps= .8*fps + .2*1/looptime #stabilizes FPS number by mostly relying on previous FPS number
    if cv2.waitKey(1) == ord('q'): #kill loop if "q" is pressed
        break
cv2.destroyAllWindows()
```

## 5. Wiring Diagram

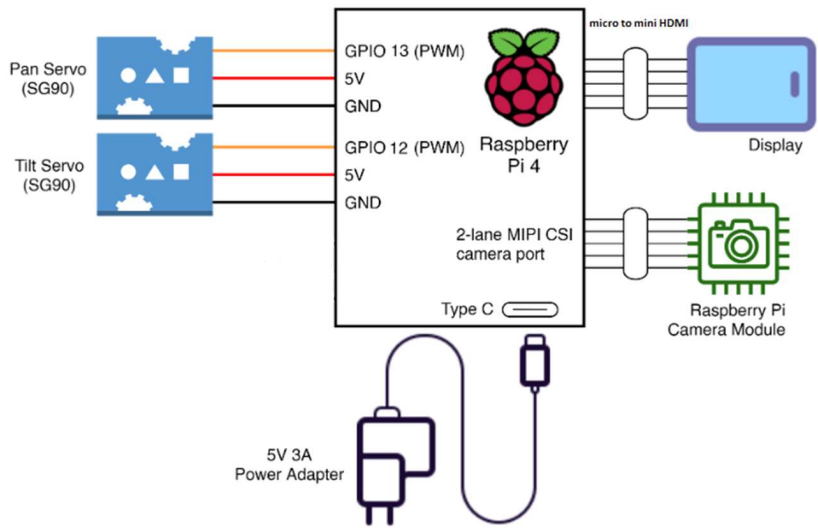

**Raspberry pi camera to Pi**

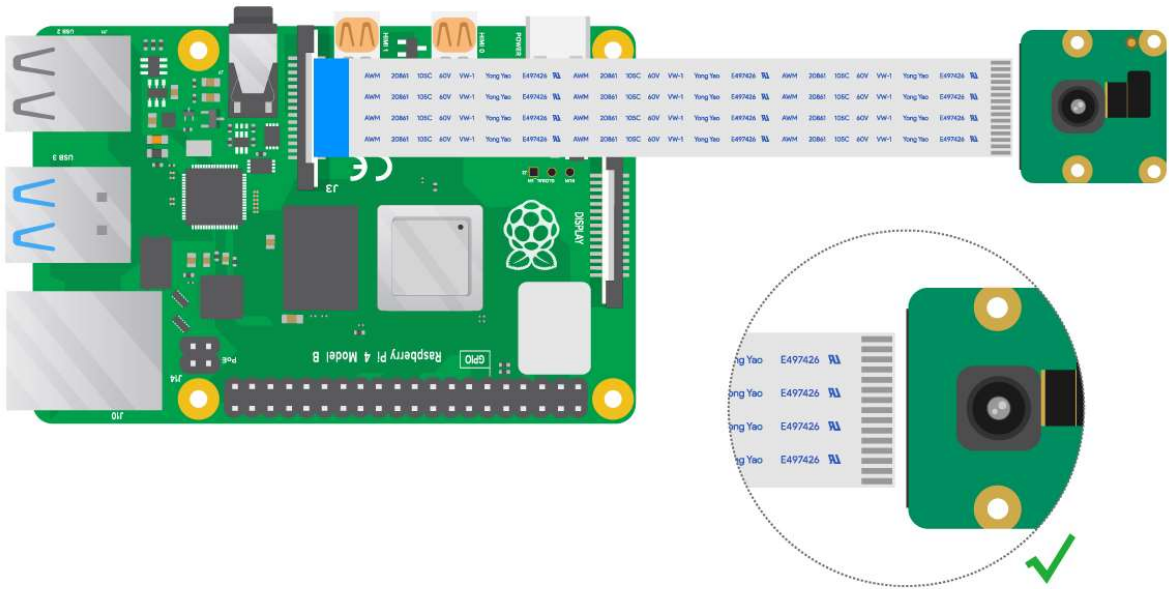

## 6. Conclusion

Today's world is evolving rapidly around us; this project was very interesting and dug deep into the surface of programming. Over the past decades there has been research on various materials and devices which help to make smaller chip used in various applications [1-14]. Yet, the world of this technology is vast even in its infancy. To master the computer language and especially the Raspberry Pi with its enormous number of configurations would take a long time to master. We believe these self-tracking cameras will be everywhere. From AI tracking security cameras to cameras that can track and keep live sporting events in the field of view will be. Camera operators will be replaced with analytical computer programmers. People who will be able to program and operate multiple systems as well as analyze the information coming in.

## References

[1] M. R. K. Akanda, "Catalogue of Potential Magnetic Topological Insulators from Materials Database", IOSR Journal of Applied Physics (IOSR-JAP) 15 (3), 22-28 (2023)

[2] M. R. K. Akanda, "Scaling of voltage controlled magnetic anisotropy based skyrmion memory and its neuromorphic application", Nano Express 10, 2 (2022). https://iopscience.iop.org/article/10.1088/2632-959X/ac6bb5/pdf

[3] Md. Rakibul Karim Akanda and Roger K. Lake, "Magnetic properties of nbsi2n4, vsi2n4, and vsi2p4 monolayers", Applied Physics Letters 119, 052402 (2021). https://doi.org/10.1063/5.0055878

[4] Md. Rakibul Karim Akanda, In Jun Park, and Roger K. Lake, "Interfacial dzyaloshinskii-moriya interaction of antiferromagnetic materials", Phys. Rev. B 102, 224414 (2020). https://journals.aps.org/prb/abstract/10.1103/PhysRevB.102.224414

[5] M. R. K. Akanda, "Catalog of magnetic topological semimetals", AIP Advances 10, 095222 (2020). https://doi.org/10.1063/5.0020096

[6] M. R. K. Akanda and Q. D. M. Khosru, "Fem model of wraparound cntfet with multi-cnt and its capacitance modeling", IEEE Transactions on Electron Devices 60, 97–102 (2013). https://ieeexplore.ieee.org/abstract/document/6375797

[7] Yousuf, A., & Akanda, M. R. K. (2023, June), *Ping Pong Robot with Dynamic Tracking* Paper presented at 2023 ASEE Annual Conference & Exposition, Baltimore, Maryland. https://peer.asee.org/43897

[8] M. R. K. Akanda and Q. D. M. Khosru, "Analysis of output transconductance of finfets incorporating quantum mechanical and temperature effects with 3d temperature distribution", ISDRS, 1–2 (2011), https://ieeexplore.ieee.org/abstract/document/6135292

[9] M. R. K. Akanda, R. Islam, and Q. D. M. Khosru, "A physically based compact model for finfets on-resistance incorporating quantum mechanical effects", ICECE 2010, 203–205 (2010). https://ieeexplore.ieee.org/abstract/document/5700663

[10] M. S. Islam and M. R. K. Akanda, "3d temperature distribution of sic mesfet using green's function", ICECE 2010,13–16 (2010). https://ieeexplore.ieee.org/abstract/document/5700541

[11] M. S. Islam, M. R. K. Akanda, S. Anwar, and A. Shahriar, "Analysis of resistances and transconductance of sic mesfet considering fabrication parameters and mobility as a function of temperature", ICECE 2010, 5–8 (2010). https://ieeexplore.ieee.org/abstract/document/5700539

[12] Md. Rakibul Karim Akanda, In Jun Park, and Roger K. Lake, "Interfacial dzyaloshinskii-moriya interaction of collinear antiferromagnets mnpt and nio on w, re, and au", APS March Meeting (2021). https://ui.adsabs.harvard.edu/abs/2021APS..MARE40004A/abstract

[13] Rakibul Karim Akanda, "3-D model of wrap around CNTEFT with multiple CNT channel and analytical modeling of its capacitances", Department of Electrical and Electronic Engineering (EEE) 2013.

[14] Akanda, Md. Rakibul Karim, "Magnetic Properties of Ferromagnetic and Antiferromagnetic Materials and Low-Dimensional Materials", University of California, Riverside ProQuest Dissertations Publishing, 2021, 28651079.